\documentclass[11pt,a4paper]{article}
\usepackage{emnlp2020}
\usepackage{times}
\usepackage{latexsym}
\usepackage{amsmath,amsthm,amssymb,pstricks,pst-node}
\usepackage{xspace,textcomp}
\usepackage{times,subcaption}
\usepackage{CJKutf8}
\usepackage{multicol}
\usepackage{multirow}

\usepackage{microtype}

\newcommand{\kld}{\mathit{KL}}

\newcommand{\vgvae}{mVGVAE\xspace}

\newcommand{\kevin}[1]{\textcolor{blue}{\bf \small [ #1 --KG]}}
\newcommand{\mingda}[1]{\textcolor{red}{\bf \small [ #1 --MC]}}
\newcommand{\sam}[1]{\textcolor{magenta}{\bf \small [ #1 --SW]}}

\renewcommand{\kevin}[1]{}
\renewcommand{\mingda}[1]{}
\renewcommand{\sam}[1]{}

\newsavebox\FrameBox
\newenvironment{Frame}{%
   \par\setbox\FrameBox\hbox\bgroup\minipage{0.45\textwidth}\parskip\baselineskip\ignorespaces
}{%
   \endminipage\egroup\fbox{\box\FrameBox}\par
}

\title{controllable paraphrase generation using bitext}

\title{Learning to Paraphrase and Translate with Syntactic Control from Parallel Text}

\title{Controllable Bilingual Paraphrase Generation with a Syntactic Exemplar}

\title{Syntactically-Controlled Paraphrase Generation and Translation from Parallel Text}

\title{Controllable Paraphrasing and Translation with a Syntactic Exemplar}
\title{Exemplar-Controllable Paraphrasing and Translation using Bitext} %

\author{Mingda Chen\qquad  Sam Wiseman\qquad Kevin Gimpel\\
Toyota Technological Institute at Chicago, Chicago, IL, 60637, USA\\
  {\tt \{mchen,swiseman,kgimpel\}@ttic.edu}\\}

\date{}

\begin{document}
\maketitle
\begin{abstract}
Most prior work on exemplar-based syntactically controlled paraphrase generation relies on automatically-constructed large-scale paraphrase datasets, which are costly to create. We sidestep this prerequisite by adapting models from prior work to be able to learn solely from bilingual text (bitext). Despite only using bitext for training, and in near zero-shot conditions, our single proposed model 
can perform four tasks: 
controlled paraphrase generation in both languages and controlled machine translation in both language directions. 
To evaluate these tasks quantitatively, we 
create three novel evaluation datasets. 
Our experimental results show that our models achieve competitive results on  controlled paraphrase generation and strong performance on controlled machine translation. Analysis shows that our models learn to disentangle semantics and syntax in their latent representations, but still suffer from semantic drift.
\footnote{Test data is available at \url{https://github.com/mingdachen/mVGVAE}.}
\end{abstract}

\section{Introduction}

\kevin{I did some editing to this paragraph and added some citations to try to bolster the motivation}\mingda{thanks!}
We consider the task of syntactically-controlled paraphrasing, which seeks to generate sentences that conform to desired syntax, specified either by syntactic templates \cite{iyyer-etal-2018-adversarial} or a sentential exemplar \cite{chen-etal-2019-controllable}. Controlled paraphrasing can be used in collaborative and assistive writing technologies, which are deployed technologies used by creative writers \citep{manjavacas-etal-2017-synthetic,miller2019artist}, 
English language learners \citep{chodorow2010utility,abbas-2020}, students in educational settings \citep{weston-sementelli-2018},  %
and marketing professionals \citep{huang-2021}. Writers use these tools to make their writing more fluent and professional, as well as to gain inspiration from seeing new ways to express their thoughts. 
Using sentential exemplars can provide a way to tailor the suggestions to match a desired syntactic pattern, without having to specify a linguistic structure such as a constituency tree. In the context of machine translation, controlled translation can be used to customize machine translation outputs in particular ways, such as producing simple sentence structures for ease of understanding. Also, controlled translation can be a way to produce multiple diverse translations which can be used for reranking or for presenting to users to provide a richer sense of the meaning of the original text when the 1-best translation is not useful \cite{mayhew-etal-2020-simultaneous}. 

Most prior work on controlled paraphrasing relies on large-scale paraphrase datasets, which are created automatically from bitext \cite{ganitkevitch-etal-2013-ppdb,wieting-gimpel-2018-paranmt,hu-aaai-2019-parabank}. However, creating paraphrase datasets is costly  \citep{wieting-gimpel-2018-paranmt,wieting-etal-2019-simple}, so in this paper we focus on ways of learning to perform exemplar-based syntactically controlled paraphrasing and translation directly from bilingual text (bitext). 
Inspired by prior work on controlled paraphrase generation \cite{chen-etal-2019-multi-task,chen-etal-2019-controllable}, we use a deep generative sentence model with two latent variables for capturing syntax and semantics. Building on the assumption that the semantics of translation pairs are shared, but syntax varies, we adapt the multi-task objectives proposed by \citet{chen-etal-2019-multi-task,chen-etal-2019-controllable}.
The model only requires training on bitext, yet it is capable of exemplar-based syntactically controlled paraphrase generation and exemplar-based syntactically controlled machine translation in an almost zero-shot manner. 

To evaluate models on these tasks quantitatively, we construct three datasets: one for Chinese controlled paraphrase generation, and the other two for Chinese-to-English and English-to-Chinese controlled machine translation. Similar to the dataset from \citet{chen-etal-2019-controllable}, each instance in these datasets contains three items: a semantic input, a syntactic exemplar, and a reference. Models must generate sentences that combine the semantics of the semantic input with the syntax of the syntactic input.  When creating these datasets, we always first automatically construct a large pool of syntactic/semantic paraphrases and then perform heavy manual post-editing to ensure the quality of the dataset and difficulty of the tasks.

Empirically, for English controlled paraphrase generation, we show that (1) our models achieve competitive results compared to prior work that uses English paraphrase pairs and stronger results compared to prior work that uses translation pairs; (2) in order for the models trained on translation pairs to reach similar performance as those trained on paraphrase pairs, the translation-trained models need more training instances than the paraphrase-trained models. We also show that our models are able to perform Chinese controlled paraphrase generation and controlled machine translation without supervision for these tasks. Quantitative analysis shows that the models learn to disentangle semantics and syntax in latent representations. Qualitative analysis shows that our models suffer from semantic drift, especially for long sentences and in the controlled machine translation tasks, showing clear directions for future research.

\section{Related Work}

Paraphrase generation uses either monolingual parallel corpora \cite{quirk-etal-2004-monolingual,prakash-etal-2016-neural,gupta2018deepparaphrase,li-etal-2018-paraphrase} or bilingual parallel corpora by pivoting \cite{bannard-callison-burch-2005-paraphrasing,ganitkevitch-etal-2013-ppdb,mallinson-etal-2017-paraphrasing}, back-translating \cite{wieting-gimpel-2018-paranmt,hu-aaai-2019-parabank,hu-etal-2019-large}, and more recently language modeling \cite{guo2019zero}. 
Our focus in this work is syntactically-controlled paraphrasing and machine translation based on bitext, whereas most prior work on controlled paraphrase generation relies on English training data \cite{iyyer-etal-2018-adversarial,chen-etal-2019-controllable,goyal2020neural,Kumar2020SyntaxguidedCG,li2020transformer,kazemnejad-etal-2020-paraphrase}. 

Recently, \citet{ijcai2020-547} show that machine translations can be used as training data for English syntactically-controlled paraphrasing.
Relatedly, \citet{wieting-etal-2019-simple} learn sentence embeddings directly from translations instead of paraphrase pairs, \citet{Wieting2019ABG} learn disentangled sentence embeddings from translations, and \citet{liu-etal-2019-exploring} learn syntactic sentence embeddings through translations and part-of-speech tags.

There is a long history of example-based machine translation \cite{nagao1984framework,somers1999example}, and recently researchers find it helpful to build generations upon sentential exemplars \cite{guu-etal-2018-generating,weston-etal-2018-retrieve}. Our approach differs from prior work in that we use sentential exemplars for controllability instead of improving generation quality. 
Part of our evaluation involves syntactically controlled machine translation, relating it to syntax-based machine translation systems \cite{wu-1997-stochastic,chiang-2005-hierarchical,sennrich-haddow-2016-linguistic}. Recently, \citet{akoury-etal-2019-syntactically} found it helpful in speeding up the decoding process.

\section{Method and Evaluation}
\subsection{Method}
To perform exemplar-based syntactically controlled generation from bitext,
we follow an approach based on the vMF-Gaussian Variational Autoencoder (VGVAE) model~\cite{chen-etal-2019-multi-task,chen-etal-2019-controllable}. VGVAE is a deep generative models with two latent variable models for modelling syntax and semantics, and it is trained with two multi-task losses designed for monolingual paraphrase sentence pairs.
\kevin{I think we need to say a bit more here in the main text about this model at a high level, in order to make the paper more self-contained. Otherwise the reviewers will express concern that the paper is not self-contained enough to understand. To free up space for this, we can just make the writing more concise in several places. I can help with this once we have all the material we want to include.}\mingda{added. what do you think?}
To adapt VGVAE and the associated multi-task losses to multilingual settings, we make a few changes, including (1) the use of byte-pair encoding (BPE; \citealp{sennrich-etal-2016-neural}); (2) we use language-specific syntactic encoders, but share the semantic encoder between the two languages; (3) we prepend a language-specific token to each input sentence. Details are in the appendix. We will call this model ``\vgvae''.

The training of the \vgvae only requires bitext, yet the trained model will contain necessary parameters to perform exemplar-based syntactically controlled paraphrase generation for either language. Moreover, training on bitext also makes the model suitable for cross-lingual tasks, such as exemplar-based syntactically controlled machine translation. In the following sections, we will demonstrate the model's ability to accomplish these tasks in an almost zero-shot fashion.

\subsection{Controllable Paraphrase Generation with a Syntactic Exemplar} 
\label{sec:paradata}

\begin{CJK*}{UTF8}{gbsn}
\begin{figure}
    \small
    \centering
    \begin{subfigure}[t]{16.55cm}
        \begin{Frame}
        $X$: {\color{red}his teammates' eyes got an ugly, hostile expression.}\\
        $Y$: {\color{blue}the smell of flowers was thick and sweet.}\\
        $Z$: the eyes of his teammates had turned ugly and hostile.
        \end{Frame}
    \end{subfigure}
    \begin{subfigure}[t]{16.55cm}
        \begin{Frame}
        $X$: {\color{red}只有团结一致，我们才能找到金子。}\\
        \phantom{$X$: }(only solidarity , we can find gold .)\\
        $Y$: {\color{blue}你如果要帮我准备早餐，就要快点起床！}\\
        \phantom{$Y$: }{\scriptsize (you if want help me prepare breakfast , then quickly get up !)} \\
        $Z$: 我们如果想找到金子，就要团结起来！ \\
        \phantom{$Z$: }(we if want find gold , then need solidarity !)
        \end{Frame}
    \end{subfigure}
    \caption{Top: an English example of exemplar-based syntactically controllable paraphrase generation from \citet{chen-etal-2019-controllable}. Bottom: a Chinese example from our annotated dataset, where each sentence is followed by a gloss. Both examples have a semantic input $X$ (red), a syntactic exemplar $Y$ (blue), and a reference output $Z$ (black). 
    }
    \vspace{-1em}
    \label{monolingual-example}
\end{figure}
\end{CJK*}

To evaluate the syntactic controllability of generation, \citet{chen-etal-2019-controllable} constructed a human-annotated dataset in English. As shown in Figure~\ref{monolingual-example}, each instance in this dataset has three sentences: a semantic input, a syntactic input, and a reference output. The semantic input shares similar semantics to the reference, but they have different syntax. The syntactic input shares similar syntax to the reference, but they have different semantics. Solving this task requires models to be able to disentangle the syntax and semantics of the input sentences and manipulate these representations to generate the desired output.

Training on bitext allows our models to perform generation in languages other than English. To evaluate this capability, we construct a similar dataset in Chinese, as shown in Figure~\ref{monolingual-example}. Since there is no large-scale Chinese paraphrase dataset available, we first automatically tag the Chinese sentences obtained from the Chinese-English OpenSubtitles corpus \cite{lison-tiedemann-2016-opensubtitles2016}, and then for each sentence, we find a sentence that shares the most similar syntax in the same corpus by computing the edit distance between the part-of-speech tag sequences. 
\kevin{I think we are supposed to include a link to \url{http://www.opensubtitles.org/} somewhere}\mingda{added it to the beginning of sec 4.1}
After obtaining these syntactic paraphrases, a native Chinese speaker paraphrases the references, following the data construction process described in \citet{chen-etal-2019-controllable} to ensure syntactic mismatch between the semantic input and the syntactic input, and semantic mismatch between the syntactic input and the reference. This contributes 378 instances. We also consider using Chinese instances that are translated from our English controlled paraphrase generation test set, and then we ask the annotator to heavily post edit these instances to ensure the criteria mentioned earlier are met. 
The final dataset contains 800 instances. 

When performing syntactically controlled generation, we first use the syntactic exemplar $x_\text{syn}$ as the input to the syntactic encoder to obtain the syntactic representation (mean of the Gaussian distribution), then we combine it with the semantic representation (mean direction of the vMF distribution) obtained from the semantic input $x_\text{sem}$ and feed them to the decoder for generation.

For this task, we report BLEU \cite{papineni-etal-2002-bleu}  and syntactic tree edit distance (ST) scores. For ST, we follow \citet{chen-etal-2019-controllable} to first parse the generated sentence and the reference with the Stanford CoreNLP toolkit~\cite{manning-etal-2014-stanford} and then compute the tree edit distance between these two parse trees. Lower ST is better. Since the ST scores between the generations and references are similar to the ones between the generations and syntactic inputs, we choose to report the former for simplicity and include the latter in the supplementary material. 
For other metrics, including ROUGE \cite{lin-2004-rouge} and METEOR \cite{banerjee-lavie-2005-meteor}, please refer to the supplementary material. 

\subsection{Controllable Machine Translation with a Syntactic Exemplar} 
\label{sec:mtdata}

\begin{CJK*}{UTF8}{gbsn}
\begin{figure}
    \small
    \centering
    \begin{subfigure}[t]{16.55cm}
        \begin{Frame}
        $W$: {\color{violet} 鹰钩鼻和僵硬的颧骨使他的脸看起来很奇怪}\\
         \phantom{$W$}{\scriptsize(hawk-bridged nose and stiff cheekbones let his face look very strange .)}\\
        $Y$: {\color{blue}clarke was a short man with bushy black brows and brown eyes.}\\
        $Z$: he had a strangely shaped face with a hawk-bridged nose and stiff cheekbones.
        \end{Frame}
    \end{subfigure}
    \begin{subfigure}[t]{16.55cm}
        \begin{Frame}
        $W$: {\color{violet}the failure of your students is the failure of yours.}\\
        $Y$: {\color{blue} \scriptsize 杰克的不负责任和布莱恩的信誓旦旦真的是天作之合。}\\
        \phantom{$Y$: }{\scriptsize(jack's irresponsibility and bryan's promises really are perfect match .)}\\
        $Z$: 你学生的失败和你的失败是一回事。\\
        \phantom{$Z$: }{\scriptsize (you student's failure and your failure are the same thing .)}
        \end{Frame}
    \end{subfigure}
    \caption{Top: an example in our Chinese-to-English syntactically controlled translation dataset. Bottom: an example in our English-to-Chinese syntactically controlled translation dataset. Both examples have semantic input $W$ (violet), syntactic exemplar $Y$ (blue), and reference output $Z$ (black). Each Chinese sentence is followed by a gloss.}
    \label{cross-lingual-example}
    \vspace{-1em}
\end{figure}
\end{CJK*}

Motivated by the fact that our models are trained on bitext, we create a novel task, where the semantic input is from the source language, and the syntactic input and reference are from the target language. Examples for both English-to-Chinese and Chinese-to-English tuples are shown in Figure~\ref{cross-lingual-example}. Compared to the task of generating paraphrases, this task is more challenging in that it requires translation and syntactic control simultaneously.

We construct evaluation datasets based on the monolingual syntactically-controlled paraphrase datasets described in the previous section. We first use Google Translate to translate the semantic inputs, and then manually go through these sentences to ensure that (1) there is no grammatical error or semantic drift in the translations; and (2) there is strong syntactic mismatch between the translations and the references, so that the task can not be trivially solved by translating the semantic inputs.

When evaluating models on these two datasets, we mostly follow the process used for paraphrase generation except that we use the language-specific syntactic encoders to encode the syntactic inputs. We report BLEU and ST scores for this task.

\section{Experiments}

\subsection{Training}
The training of our models uses bitext from OpenSubtitles.\footnote{\url{http://www.opensubtitles.org/}} We report results on Chinese-English (Zh-En) sentence pairs.
Results on German-English (De-En), Spanish-English (Es-En), and French-English (Fr-En) sentence pairs are in the supplementary material.
As baselines, we train models on Czech-English (Cs-En) sentence pairs from the CzEng corpus \cite{czeng16:2016} and English-English (En-En) sentence pairs from the ParaNMT-50M dataset \cite{wieting-gimpel-2018-paranmt}, which is constructed by back-translating the CzEng corpus.

For all these datasets, we first tokenize text with the Stanford CoreNLP toolkit and then use BPE with 30,000 merge operations.  \citet{chen-etal-2019-controllable} used several heuristics for additional filtering of the 5-million-pair preprocessed subset of ParaNMT-50M released by \citet{wieting-gimpel-2018-paranmt}, eventually using half a million paraphrase pairs for training.  
As these heuristics are not directly applicable to bitext, we use the full 5-million preprocessed subset for En-En. For Zh-En, we randomly sample 5 million sentence pairs. To make the results more comparable to prior work on English controllable paraphrasing, for both settings and the CzEng corpus, we also report results that use 0.5 million sentence pairs, randomly sampled.
We use word noising during training \cite{chen-etal-2019-controllable}, and the probability of noising a word is 90\%. We perform early stopping based on the BLEU score from the development set.
More details, such as runtime and hyperparameters, are in the appendix. 

\kevin{I added the following text. I think this is good to state before we report results. I'm not sure if this is the best place for it and I wasn't sure if it was stated somewhere before this though.}\mingda{thanks!}
We only have a development set for the English paraphrase generation task, for which we use the dev split from \citet{chen-etal-2019-controllable}. For the other tasks, we do not have development sets. We treat our newly-created datasets (described in Sections \ref{sec:paradata} and \ref{sec:mtdata}) as test sets and report results on them. Therefore, we can consider our results on these tasks to be zero-shot or zero-shot crosslingual results.

\subsection{Results}

\paragraph{Controlled Paraphrase Generation.} For  English and Chinese paraphrase generation, we report two groups of baselines. The first group is ``return-input'', where we return either the syntactic input or the semantic input as the prediction. As shown in Tables \ref{tab:en-para-gen} and \ref{tab:zh-para-gen}, compared to the semantic input, the syntactic input leads to better ST score but worse semantic-related scores, i.e., BLEU.\footnote{When evaluating Chinese sentences, the metrics are computed at the character-level, following the process described in~\citet{ma-etal-2019-results}.} This reflects our consideration about the differences between these two when constructing the dataset. 

For English paraphrasing, the second group of baselines is models trained on ParaNMT. The empirical comparison to prior work \citep{chen-etal-2019-controllable} shows the impact of several changes made when modifying \vgvae for use with bitext, such as BPE, as well as the difference in training data filtering. Our model (En-En) achieves worse performance in semantic-related metrics compared to prior work when training on the same amount of data. Using more data partially mitigates the performance gap.

\begin{table}[t]
    \centering
    \setlength{\tabcolsep}{4pt}
    \small
\begin{tabular}{|l|c|c|}\hline
& \%BLEU ($\uparrow$) & ST ($\downarrow$) \\\hline
\multicolumn{3}{|c|}{Return-input baselines}\\\hline
Syntactic input & 3.3  & 5.9 \\
Semantic input & 18.5  & 12.0 \\\hline
\multicolumn{3}{|c|}{Models trained on the ParaNMT dataset}\\\hline
\citet{chen-etal-2019-controllable} (0.5M En-En) 
& 13.6 & 6.7 \\
0.5M En-En & 11.7 & 6.8 \\
5M En-En & 13.0 & 6.6 \\\hline
\multicolumn{3}{|c|}{Models trained on bitext}\\\hline
\citet{ijcai2020-547} (0.5M Zh-En) & 8.7 & 9.8 \\
0.5M Cs-En & 9.3 & 7.6 \\
0.5M Zh-En & 9.9 & 7.2 \\
5M Zh-En & 11.5 & 6.7 \\
5M Zh-En Big & 12.5 & 6.7 \\\hline
\end{tabular}
    \caption{English controlled paraphrase generation.}
    \vspace{-0.5em}
    \label{tab:en-para-gen}
\end{table}

For models trained on bitext, our model obtains stronger performance than \citet{ijcai2020-547} when training on the same amount of data. However, the results for our models and \citet{ijcai2020-547} are not strictly comparable as their models are trained on different datasets.
The difference between the results for 0.5M En-En and 0.5M Cs-En shows the advantage of learning from monolingual paraphrase pairs as compared to bitext. %
Compared to training on 0.5M Cs-En, we are able to obtain slightly better performance when training on 0.5M Zh-En sentence pairs.\footnote{We also experimented with other language pairs in order to study the relationship between performance and language choice but did not observe any clear patterns. More details are in the supplementary material.} 
We also find that after increasing the amount of training data to 5M, we are able to match the performance of the 0.5M En-En model. 
Prior work on learning sentence representations from bitext \cite{wieting-etal-2019-simple} also found that training on bitext requires more data than training on English paraphrase data. In light of the strong performance, we will report results that use 5M sentence pairs for the following experiments.

Although models trained on 5M Zh-En bitext perform worse on English paraphrase generation in terms of the BLEU score, they show strong performance on the ST score, suggesting that these generations share similar syntax with the syntactic inputs but they are not faithful to the semantic inputs (e.g., in Sec. \ref{sec:gen_sample}, we find that the model suffers from semantic drift).
It is also worth noting that (1) compared to the model trained on ParaNMT, the model trained on Zh-En only trains on half the number of English sentences; (2) during training, the models never get to train on paired monolingual sentences, yet they manage to control the syntax in the generations. 

Nonetheless, to offer a competitive baseline for future work, we train a large model with 1000 hidden units per direction on 5 million instances of Zh-En bitext, and report the results as ``Zh-En Big'' in the tables. As shown in Tables \ref{tab:en-para-gen} and \ref{tab:zh-para-gen}, increasing model size boosts performance significantly.

\begin{table}[t]
    \centering
    \setlength{\tabcolsep}{5pt}
    \small
\begin{tabular}{|l|c|c|}\hline
& \%BLEU ($\uparrow$) & ST ($\downarrow$) \\\hline
\multicolumn{3}{|c|}{Return-input baselines}\\\hline
Syntactic input & 6.2 & 13.9 \\
Semantic input & 49.0 & 18.7 \\\hline
\multicolumn{3}{|c|}{Our work}\\\hline
Zh-En & 12.8 & 15.8 \\
Zh-En Big & 16.6 & 15.5 \\\hline
\end{tabular}
    \caption{Chinese controlled paraphrase generation.}
    \vspace{-0.7em}
    \label{tab:zh-para-gen}
\end{table}

\begin{table}[t]
    \centering
    \setlength{\tabcolsep}{5pt}
    \small
\begin{tabular}{|l|c|c|}\hline
& \%BLEU ($\uparrow$) & ST ($\downarrow$) \\\hline
\multicolumn{3}{|c|}{Return-input baselines}\\\hline
Syntactic input & 3.3 & 5.9 \\\hline
\multicolumn{3}{|c|}{Neural machine translation baselines}\\\hline
OpenNMT & 11.0 & 11.5 \\
Google Translate & 14.5 & 11.6 \\\hline
\multicolumn{3}{|c|}{Our work}\\\hline
Zh-En & 10.9 & 6.7 \\
Zh-En Big & 12.1 & 6.6 \\\hline
\end{tabular}
    \caption{%
    Chinese$\rightarrow$English controlled translation.}
    \vspace{-0.5em}
    \label{tab:zh2en-para-gen}
\end{table}

\begin{table}[t]
    \centering
    \setlength{\tabcolsep}{5pt}
    \small
\begin{tabular}{|l|c|c|}\hline
& \%BLEU ($\uparrow$) & ST ($\downarrow$) \\\hline
\multicolumn{3}{|c|}{Return-input baselines}\\\hline
Syntactic input & 6.2 & 13.9 \\\hline
\multicolumn{3}{|c|}{Neural machine translation baselines}\\\hline
OpenNMT & 12.3 & 19.2 \\
Google Translate & 30.3 & 18.5  \\\hline
\multicolumn{3}{|c|}{Our work}\\\hline
Zh-En & 9.6 & 16.1 \\
Zh-En Big & 11.9 & 15.7  \\\hline
\end{tabular}
    \caption{%
    English$\rightarrow$Chinese controlled translation.}
    \vspace{-0.5em}
    \label{tab:en2zh-para-gen}
\end{table}

\paragraph{Controlled Machine Translation.} As we do not have semantic inputs in the same language as references in this setting, the ``return-input'' baseline for the semantic input would score very badly. So, we evaluate standard machine translation systems on these tasks as baselines, simply applying them to the semantic input and ignoring the syntactic input. The systems we consider include (1) a neural sequence-to-sequence model trained on the same 5 million Zh-En bitext using  OpenNMT~\cite{klein-etal-2017-opennmt}\footnote{We use a 2-layer bidirectional LSTM on the encoder and a 2-layer unidirectional LSTM on the decoder. Both LSTMs have 500 hidden units per direction.}; and (2) Google Translate. 

Test results are shown in Tables \ref{tab:zh2en-para-gen} and \ref{tab:en2zh-para-gen}. We note that to obtain the OpenNMT results in these two tables, we need to train two separate models on two directions of the Zh-En bitext, whereas we can use a single \vgvae model for both translation directions (in addition to both paraphrasing tasks). The two tables show that these machine translation systems achieve strong performance on the semantic metric (BLEU), but weak performance on the syntactic metric (ST). This highlights that success of syntactic controllability of generation depends on using the syntactic inputs. Our large model achieves the best results in ST in these two tables and still manages to obtain reasonable BLEU scores.

\section{Analysis}

\begin{table}[t]
\small
    \centering
\begin{tabular}{|l|c|c|c|}\hline
\multirow{2}{*}{ } & \multicolumn{3}{c|}{ English test set } \\\cline{2-4}
& sem. & syn. & $\Delta$ \\\hline
Prior work (En-En) & 74.3 & 7.4 & 66.9 \\
Our work (En-En) & 74.9 & 12.0 & 62.9 \\
Our work (Zh-En) & 73.6 & 16.3 & 57.3 \\\hline
& \multicolumn{3}{c|}{ Chinese test set } \\\hline
fastText & \multicolumn{3}{c|}{ 74.0 } \\\hline
Our work & 76.2 & 23.5 & 52.8 \\\hline
\end{tabular}
    \caption{Pearson's correlation (\%) for the semantic similarity tasks.}
    \vspace{-1em}
    \label{tab:sts-res}
\end{table}

\begin{table}[t]
    \centering
    \setlength{\tabcolsep}{5pt}
    \small
\begin{tabular}{|l|c|c|c|c|c|c|}
\hline
\multirow{2}{*}{ } & \multicolumn{3}{c|}{ POS Accuracy } & \multicolumn{3}{c|}{ CP Labeled $F_1$ } \\\cline{2-7}
& sem. & syn. & $\Delta$ & sem. & syn. & $\Delta$ \\\hline
& \multicolumn{6}{|c|}{ English test set } \\\hline
Oracle & \multicolumn{3}{c|}{62.3} & \multicolumn{3}{c|}{71.1} \\
Random & \multicolumn{3}{c|}{12.9} & \multicolumn{3}{c|}{19.2} \\\hline
PW (En-En) & 20.3 & 43.7 & 23.4 & 24.8 & 40.9 & 16.1 \\\hline
En-En & 19.6 & 44.9 & 25.3 & 24.2 & 42.4 & 18.2 \\\hline
Zh-En & 19.7 & 44.4 & 24.7 & 24.4 & 41.7 & 17.3 \\\hline
& \multicolumn{6}{|c|}{ Chinese test set} \\\hline
Oracle & \multicolumn{3}{c|}{ 56.6 } & \multicolumn{3}{c|}{ 58.2 } \\
Random & \multicolumn{3}{c|}{ 13.4 } & \multicolumn{3}{c|}{ 17.8 } \\\hline
fastText & \multicolumn{3}{c|}{ 18.7 } & \multicolumn{3}{c|}{ 21.7 } \\\hline
Zh-En & 15.0 & 36.2 & 21.2 & 19.3 & 32.3 & 12.9 \\\hline
\end{tabular}
    \caption{POS tagging accuracy (\%) and constituent parsing labeled $F_1$ scores (\%) for the syntactic evaluation. PW = prior work.}
    \vspace{-1em}
    \label{tab:syn-res}
\end{table}

\begin{CJK*}{UTF8}{gbsn}
\begin{table*}[t]
    \centering
    \small
\begin{tabular}{|p{0.3\textwidth}|p{0.3\textwidth}|p{0.31\textwidth}|}\hline
\multicolumn{1}{|c|}{Query Sentence} & \multicolumn{1}{|c|}{Semantically Similar} & \multicolumn{1}{|c|}{Syntactically Similar} \\\hline
with a gun , you feel more dangerous . i would n't know . & i 'll give him a gun if it makes you feel any better . & all these days , i could n't work , i could n't sleep . \\\hline
你 从 窗户 外 可以 看到 它 或在 电视 上 也 会 看到 它 & 你 几乎 不 可能 再 近 距离 看到 它的 ， 用 望远镜 也 不行 & 我 在 国会 上 讲 过 我们 需要 建桥 但 我们 会 靠 自己 \\
(you from window outside can see it or in tv also can see it) & (you almost no possible again close distance see it , use telescope also not work) & (i at congress talked we need build a bridge but we can depend on ourselves) \\\hline
\end{tabular}
    \caption{The most similar sentences to particular query sentences in terms of the semantic variable and syntactic variable based on the model trained on Zh-En bitext. Each Chinese sentence is followed by a gloss. More examples are in the supplementary material.}
    \vspace{-1em}
    \label{tab:near-neighbor-analysis-mono}
\end{table*}
\end{CJK*}

\subsection{Human Evaluation}
\begin{table}[t]
    \centering
    \small
    \begin{tabular}{|l|c|c|}
    \hline
        & Avg. & Std. \\\hline
        OpenNMT & 2.4 &  1.2 \\
        Google Translate & 2.4 & 1.1 \\
        Our model (Zh-En Big) & 3.5 & 1.4\\\hline
    \end{tabular}
    \caption{Human evaluation results for Chinese-to-English controlled machine translation. We report the average and standard deviation of scores that measures the syntactic similarity between the generations and the syntactic inputs. Higher is better for the average scores.}
    \vspace{-0.5em}
    \label{tab:human_evaluation}
\end{table}

We conduct a human evaluation to verify the extent to which the syntax of generations matches the syntax of the syntactic inputs. We use Amazon Mechanical Turk and ask human annotators to give scores ranging from 1 to 5 (with 1 being the most dissimilar) for the syntactic similarity between the generated output and the syntactic inputs (details %
are in the supplementary material). 
We collect 79 annotations (1 annotation per instance) for each of three systems. Similar to what we observe with the ST metric, our model performs the best on the average score among the three systems (Table \ref{tab:human_evaluation}). These results provide validation that our ST metric captures syntactic similarity as measured by human judges.

\subsection{Disentanglement of Latent Representations}
To evaluate the extent of disentanglement of the syntactic and semantic representations, we follow the idea of \citet{chen-etal-2019-multi-task} and test our trained models on both semantic and syntactic similarity tasks. When computing sentence representations, we first use the inference model $q_\phi(y\vert x)$ or $q_\phi(z\vert x)$ and then we either use the mean direction of vMF distribution or the mean of Gaussian distribution to obtain the semantic or syntactic representations for sentences. 
We expect that compared to the syntactic latent representations, the semantic ones would perform better on semantic tasks and worse on syntactic tasks, and vice versa. The performance gap between the two representations can be used as an indicator for the disentanglement of syntactic variables and semantic variables.

For the semantic evaluation of English sentences, we use the test set from the sentence textual similarity (STS) benchmark~\cite{cer-etal-2017-semeval}. For Chinese, we use the 1360 human-annotated instances from \citet{wang-etal-2017-exploiting}. These datasets provide human annotations for pairs of sentences, indicating the semantic similarity of the given sentence pair. We report the Pearson correlation between the human annotation and the cosine similarity computed based on the sentence representations.

For the syntactic evaluation, we follow the procedure described by \citet{chen-etal-2019-multi-task} to automatically parse and tag sentences using the Stanford CoreNLP toolkit for all languages except English. 
For English, we use the existing dataset provided by \citet{chen-etal-2019-multi-task}. 
Then, we randomly pick 300 sentences for each length (up to 30) as test sets, and leave the rest of the sentences with the same length as candidates. We use the sentence representations to retrieve the nearest neighbor in the candidate pool for sentences in the test set, and compute the distance metrics between these two sentences by computing labeled $F_1$ scores for constituency parse (CP) trees or accuracies for part-of-speech (POS) tagging. The syntactic match between the nearest neighbors and the query sentences can illustrate the extent of syntactic information that the sentence representations have captured. This kind of retrieval-based approach has been shown to be effective in sequence labeling~\cite{wiseman-stratos-2019-label}.

\begin{table*}[t]
\small
    \centering
\begin{tabular}{|p{0.22\textwidth}|p{0.22\textwidth}|p{0.22\textwidth}|p{0.22\textwidth}|}\hline
 \multicolumn{1}{|c|}{Semantic Input} & \multicolumn{1}{c|}{Syntactic Input}  & \multicolumn{1}{c|}{En-En Model} & \multicolumn{1}{c|}{Zh-En Big Model} \\\hline

this 's a crime , you know that , right ? &  they understand that was a chance ? & you know that was a crime ? & you know this was a crime ? \\\hline

how exciting that is . & that 's exactly right . & that 's so exciting . & that 's so exciting . \\ \hline

the mayor will be called for responsibility by citizens when these policies fail . & when this report is found , it will give us ideas on the future of the factory .   & when the government is informed , they will call it the citizens of the responsibility of policy . & when the mayor gets hurt , they will call you a lot of citizens for a while . \\\hline
by his side crouched a huge black wolfish dog . & a giant yellow bird lives on it . & the black fish wolf catches behind him . & a fish black market is beside him .  \\\hline

\end{tabular}

    \caption{Example outputs from the ``En-En'' and ``Zh-En Big'' models for the English controlled paraphrase generation tasks. Both models are trained on 5 million sentence pairs.}
    \label{tab:eng-comparison-examples}
\end{table*}

\begin{CJK*}{UTF8}{gbsn}
\begin{table*}[t]
\small
    \centering
\begin{tabular}{|p{0.30\textwidth}|p{0.30\textwidth}|p{0.32\textwidth}|}\hline
 \multicolumn{1}{|c|}{Semantic Input} & \multicolumn{1}{c|}{Syntactic Input}  & \multicolumn{1}{c|}{Generation} \\\hline
\multicolumn{3}{c}{\bf Chinese Controlled Paraphrase Generation } \\\hline
我 厌倦 了 你 的 深夜 探访 。 (i tired of your late night vists.) & 他们 的 耽搁 使 他 发疯 了 。 (their delay made him crazy .) & 你 的 访客 让 我 厌烦 了 。 (your visitor made me tired .) \\\hline

\multicolumn{3}{c}{\bf English-to-Chinese Controlled Machine Translation } \\\hline

They succeeded because they found the same rhythm & 因为 它 呈现 了 最下 乘 的 科学 ， 所以 我们 不 应该 使用 它 。 (because it shows the worst science , so we should not use it . ) & 因为 他们 发现 了 最 成功 的 节奏 ， 因为 他们 都 找到 了 它 。 (because they found the most successful thythm , because they all found it .) \\\hline

\multicolumn{3}{c}{\bf Chinese-to-English Controlled Machine Translation } \\\hline
我 不 能 成为 一 名 吸毒者 或 酗酒者 ？ 为什么 ？ (i can not become a drug addict or an alcoholic ? why ?) & why ca n't it eat dried peas or something ? & why could n't i have been drinking or something ? \\\hline
\end{tabular}

    \caption{Example outputs from the ``Zh-En Big'' model for our three new controlled generation tasks. Each Chinese sentence is followed by a gloss. More examples are in the supplementary material.}
    \vspace{-1em}
    \label{tab:generation-examples}
\end{table*}
\end{CJK*}

\paragraph{Semantic Evaluation.} We report results for the English semantic similarity task in Table~\ref{tab:sts-res}. We report the best model from \citet{chen-etal-2019-controllable} as ``Prior work''. For the Chinese test set, we also benchmark fastText \cite{bojanowski-etal-2017-enriching} on it by using the averaged word vectors as the sentence representation. In comparing to prior work, our En-En model performs worse on the $\Delta$ value, and performs better on the semantic variable.

In general, as suggested by the large $\Delta$ values, our models learn to disentangle the semantic information in the syntactic variable and semantic variable, offering results competitive with prior work. This shows that training on bitext yields similar disentanglement as training on paraphrases. 
This phenomena also generalizes to other languages, as shown in Table~\ref{tab:sts-res}, where we report results for the Chinese semantic similarity task (see the supplementary material for more results on other languages). Although the semantic variable outperforms fastText by around 2 points, the syntactic variable has a much lower result, leading to a $\Delta$ value similar to the English test set. It is worth noting that when computing semantic representations, there is no language-specific parameter aside from special language tokens that are prepended to the input sequences, yet the models still manage to learn to encode semantic information for different languages.

\paragraph{Syntactic Evaluation.} We report results for the English syntactic evaluation and the Chinese syntactic evaluation in Table~\ref{tab:syn-res}. We also report two baseline results for the datasets. One is ``Oracle'', where we use the parse trees or POS tags to search for nearest neighbors. This can serve as upper bound performance of the task. The other one is ``Random'', where we randomly select sentences as our predictions, and we report results averaged over ten runs. This can serve as a lower bound performance of the task. 
For the Chinese test set, we also report a fastText baseline based on the averaged word vectors.

Similar to what we observe in the semantic similarity tasks, our models can disentangle syntactic information in the latent variables, either in the English dataset or the datasets in other languages (see supplementary materials for more results on other languages). On the Chinese test set, fastText is slightly better than the semantic variable of our model, whereas the syntactic variable achieves much better performance.

\subsection{Nearest Neighbors}

We analyze nearest neighbors based on cosine similarities between query sentences drawn from the test sets and candidates with the same length from the syntactic evaluation candidate set. 
Table~\ref{tab:near-neighbor-analysis-mono} shows the nearest neighbors found by either the semantic variable or the syntactic variable. Similar to \citet{chen-etal-2019-controllable}, who trained on paraphrases, models trained on bitext also capture different characteristics for different latent variables. The semantically similar sentences share similar topics with the query sentences, while their syntactic structures are very different. The syntactically similar sentences have similar syntax to the query sentences, while their topics are different. For example, ``with a gun ... wouldn't know.'' and ``i'll give him ... any better'' both talk about guns, whereas ``all these days ... couldn't sleep'' is about working and sleeping, which is topically unrelated to the previous two sentences. However, the syntactic variable still gives it the highest similarity due to the similar syntactic structure. Similar effects can be observed for the Chinese nearest neighbors.

\subsection{Generation Samples}
\label{sec:gen_sample}

In Table \ref{tab:eng-comparison-examples}, we compare generation examples from the ``En-En'' model to those from the ``Zh-En Big'' model. Both models are trained on 5 million sentence pairs. In general, we find the models are able to generate sentences that exhibit the expected syntax without changing the semantics. However, for long inputs, the outputs from the Zh-En model become more noisy in terms of semantic preservation. 
For example, see the third instance in the table, where the En-En output is a bit garbled but still uses topically-relevant phrases (e.g., ``responsibility of policy'') whereas the Zh-En output has changed the semantics dramatically (``the mayor gets hurt''). In addition, we find that the subword tokenization may have a negative impact on the model performance. In the last example in the table, although the semantic and the syntactic inputs never talk about fish, the outputs from both models contain the word ``fish''\kevin{I think you meant ``fish''?}\mingda{updated}, which is likely due to the fact that the ``wolfish'' in the semantic input is tokenized into ``wol@@ fish'' by the byte pair encoding where ``@@'' is the separator between non-final subword units.

In Table \ref{tab:generation-examples}, we demonstrate generation examples from the Zh-En Big model for the other three tasks considered in this paper. In general, the model tends to give higher priority to the syntactic input than the semantic input, even at the cost of faithfulness to the semantic input. For example, in Chinese controlled paraphrase generation, the semantic input ``i tired of your late night visits'' gets transformed into ``your visitor made me tired.'', which shows strong syntactic similarity to the syntactic input, although the word ``visit'' is replaced by the word ``visitor''. This sort of  semantic drift occurs more often in controlled machine translation. For example, in English-to-Chinese translation, the instance with similar semantic and syntactic input is translated into ``your visitor made me late'', which mistranslates both ``visit'' and ``tired'', although the syntax of the instance remains very similar to that of the syntactic input. 

However, it is worth noting that the model manages to transform function words into appropriate words that fit the context. For example,
in Chinese-to-English controlled machine translation, the phrase ``why can't it eat'' is transformed into ``why couldn't i have been''. It is also interesting to see that the model only needs to be trained once on Zh-En bitext to perform all four of these tasks.

\section{Conclusion}
We tailored a disentanglement method so that it can be trained on bitext instead of paraphrases, and demonstrated that it can learn to perform exemplar-based syntactically controlled paraphrasing and machine translation in an almost zero-shot fashion. We annotated three new datasets targeting these tasks which will be made available. 
Quantitative analysis shows that our model learns to disentangle semantics and syntax in the latent representations, though it still suffers from semantic drift when performing controlled generation, suggesting directions for future work.

\bibliographystyle{acl_natbib}
\bibliography{anthology,emnlp2020}

\appendix

\section{Details of Methods}
We begin by briefly reviewing the learning objective and parameterization of \vgvae, and then describe the changes we make to adapt it to the multilingual setting. For more details, please refer to the original papers.

\paragraph{Learning Objectives.} \vgvae is a neural latent-variable sentence model with two latent variables, one for modeling semantics (denoted by $y$ and drawn from a von Mises-Fisher prior) and one for syntax (denoted by $z$ and drawn from a Gaussian prior).

\vgvae assumes sentences $x$ are generated by independent latent variables $y$ and $z$, leading to a factorized joint probability $p_{\theta}(x, y, z) = p_{\theta}(y) p_{\theta}(z) p_{\theta}(x \, | \, y, z)$. Furthermore, \vgvae assumes a factorized approximated posterior $q_\phi(y,z\vert x)=q_\phi(y\vert x)q_\phi(z\vert x)$. 
Combining these two assumptions gives rise to one of the learning of objectives of \vgvae, the ELBO:
\begin{equation}
\begin{aligned}
    &\log p_\theta(x)\geq\mathop\mathbb{E}_{\substack{y\sim q_\phi(y\vert x)\\z\sim q_\phi(z\vert x)}}[\log p_\theta(x\vert z,y)]\\&-\kld(q_\phi(z\vert x)\Vert p_\theta(z))-\kld(q_\phi(y\vert x)\Vert p_\theta(y))
\end{aligned}
\label{eq:elbo}
\end{equation}
Similar to \citet{chen-etal-2019-controllable}, we associate weights with the KL terms. In our experiments, we use $1e^{-3}$ for $z$ and $1e^{-4}$ for $y$. To adapt VGVAE to bitext, we use subword units, i.e., byte-pair encoding (BPE; \citealp{sennrich-etal-2016-neural}), instead of whole word tokenization.

In addition to Eq.~(\ref{eq:elbo}), \citet{chen-etal-2019-controllable} introduced other   losses for multi-task training: a paraphrase reconstruction loss (PRL) and a word position loss (WPL). We directly apply these two losses to our bitext setting. In particular, PRL for a parallel sentence pair ($x_1$,$x_2$) takes the form
\begin{equation}
\begin{aligned}
    \mathop\mathbb{E}_{\substack{y_2\sim q_\phi(y\vert x_2)\\z_1\sim q_\phi(z\vert x_1)}}[&\log p_\theta(x_1\vert y_2,z_1)] +\\ \mathop\mathbb{E}_{\substack{y_1\sim q_\phi(y\vert x_1)\\z_2\sim q_\phi(z\vert x_2)}}[&\log p_\theta(x_2\vert y_1,z_2)]
\end{aligned}
\label{eqn:prl}
\end{equation}
\noindent That is, instead of reconstructing a sentence from its syntactic variable and the semantic variable of its paraphrase, we reconstruct a sentence from its syntactic variable and the semantic variable of its \emph{translation}. 

The WPL loss adds a classifier to explicitly predict word position at each time step $t$ using the concatenation of subword unit embedding $e_t$ and the syntactic variable $z$ produced by the approximated posterior. Formally, WPL is defined as follows:
\begin{equation}
    \mathop\mathbb{E}_{z\sim q_\phi(z\vert x)}\left[\sum_{t}\log\textrm{softmax}(f([e_t;z]))_{w_t}\right]
    \nonumber
\end{equation}
\noindent where $f$ is a 3-layer feedforward neural network, $\textrm{softmax}(\cdot)_{i}$ indicates the probability at position $i$, and $w_t$ is the word boundary at $t$, i.e., the position in $x$ of the original word that contains the $t$-th subword unit. 
Though we use BPE encoding, we define the ground truth positions for WPL using the original word boundaries. Unlike \citet{chen-etal-2019-controllable}, we find the latent code does not help model performance when using subword units, so we do not include it in our models.

Finally, the learning objective for \vgvae is
\begin{equation}
    \mathit{ELBO} + \mathit{PRL} + \mathit{WPL}
\end{equation}

\paragraph{Parameterization.} Similar to~\citet{chen-etal-2019-controllable}, we parameterize the syntactic encoder $q_\phi(z\vert x)$ with a bidirectional long short-term memory~(LSTM;~\citealp{hochreiter1997long}) network, and the semantic encoder $q_\phi(y\vert x)$ with a word averaging module. Each is then followed by a 2-layer feedforward neural network for encoding the mean direction of the vMF distribution or the mean and the variance of the Gaussian distribution. The $p_\theta(x\vert z, y)$ is parameterized with a unidirectional LSTM.

We also allocate language-specific parameters to better model bitext. To indicate the languages of the input sentences, we prepend a language-specific token to each input sentence. Additionally, we use language-specific syntactic encoders, but share the semantic encoder between the two languages as we assume that the semantics between sentence pairs are shared. The syntactic and semantic encoders do not share subword unit embeddings. 

The training of the \vgvae only requires bitext, yet the trained model will contain necessary parameters to perform exemplar-based syntactically controlled paraphrase generation for either language. Moreover, training on bitext also makes the model suitable for cross-lingual tasks, such as exemplar-based syntactically controlled machine translation. In the following sections, we will demonstrate the model's ability to accomplish these tasks in an almost zero-shot fashion.

\section{Hyperparameter and Model Size}
We follow the hyperparameters used in \citet{chen-etal-2019-controllable} and did not perform any other hyperparameter search for fair comparison. We use 50 as the latent dimension for most of the models, except for the big model, where we use 100. Similarly, we use beam size 10 when evaluating models on test sets.

Our big model has 220.5 million parameters, and the other mVGVAE models each has 30.5 million parameters. OpenNMT baseline has 47.4 million parameters.

\section{Generation Samples}

\begin{CJK*}{UTF8}{gbsn}
\begin{table*}[t]
\small
    \centering
\begin{tabular}{|p{0.30\textwidth}|p{0.30\textwidth}|p{0.30\textwidth}|}\hline
 \multicolumn{1}{|c|}{Semantic Input} & \multicolumn{1}{c|}{Syntactic Input}  & \multicolumn{1}{c|}{Generation} \\\hline
\multicolumn{3}{c}{\bf Chinese Controlled Paraphrase Generation } \\\hline
我 厌倦 了 你 的 深夜 探访 。 (i tired of your late night vists.) & 他们 的 耽搁 使 他 发疯 了 。 (their delay made him crazy .) & 你 的 访客 让 我 厌烦 了 。 (your visitor made me tired .) \\\hline
我 认为 ， 如果 她 愿意 选择 这个 坟墓 ， 我们 没有 理由 将 她 嫁给 一 个 贵族 。 (i think , if she willing choose this grave , we no reason will she marry a noble ) & 如果 那 是 他们 想要 的 ， 你们 为什么 要 这样 玩 呢 ？ (if that is they want , you why like this play ?) & 如果 她 是 我们 选择 的 ， 她 就 不 会 结婚 了 。 (if she is we choose , she will not marry .)  \\\hline
帮助 我们 学习 是 布拉德 的 目的 (help us learn is brad's purpose) & 瑞德 是 去 坑害 那些 百姓 的 。 (reed is going entrap those common people .) & 布拉德 是 想 帮助 我们 学习 的 。 (brad is want help us learn .)\\\hline

\multicolumn{3}{c}{\bf English Controlled Paraphrased Generation } \\\hline

no demonic force can harm you while this blade is at your hip . & as soon as i start the escape sequence , an alarm will sound . & as long as you 're a little devil , no one will hurt .  \\\hline
why are n't his motives written down ? & why not take a look at your verdict ? & why not write the letter of his motive ?  \\\hline
no more drugs for him ! & do n't watch it all day ! & do n't give him any drugs ! \\\hline

\multicolumn{3}{c}{\bf English-to-Chinese Controlled Machine Translation } \\\hline

Unless nothing happened to him , why would n't we know ? & 如果 是 笑着 说话 ， 別人 是 能 听 出来 的 。 (if is smile talk , other people can hear it . ) & 如果 是 想知道 原因 ， 事情 会 是 不 会 发生 的 ？ (if is want to know the reason , the thing is not going to happen ?) \\\hline
I 'm tired of your late night visits . & 他们 的 耽搁 使 他 发疯 了 。 (their delay made him crazy .) & 你 的 访客 让 我 迟到 了 (your visitor made me late .) \\\hline
They succeeded because they found the same rhythm & 因为 它 呈现 了 最下 乘 的 科学 ， 所以 我们 不 应该 使用 它 。 (because it shows the worst science , so we should not use it . ) & 因为 他们 发现 了 最 成功 的 节奏 ， 因为 他们 都 找到 了 它 。 (because they found the most successful thythm , because they all found it .) \\\hline

\multicolumn{3}{c}{\bf Chinese-to-English Controlled Machine Translation } \\\hline
你 知道 你 已经 是 个 死人 吗 ？ (you know you already is a dead man ?) & do you need them stopped immediately ? & do you know you were dead ? \\\hline
我 不 能 成为 一 名 吸毒者 或 酗酒者 ？ 为什么 ？ (i can not become a drug addict or an alcoholic ? why ?) & why ca n't it eat dried peas or something ? & why could n't i have been drinking or something ? \\\hline
给 史密斯 先生 买些 面包 。 (give mr. smith buy some bread .) & let 's take a look at the mechanism . & let 's get some money for a drink . \\\hline
\end{tabular}

    \caption{Example outputs for our four controlled generation tasks. Each Chinese sentence is followed by a gloss.}
    \vspace{-1em}
    \label{tab:generation-examples2}
\end{table*}
\end{CJK*}

We show generation examples in Table \ref{tab:generation-examples2}.

\section{Nearest Neighbors}

\begin{CJK*}{UTF8}{gbsn}
\begin{table*}[t]
    \centering
    \small
\begin{tabular}{|p{0.3\textwidth}|p{0.3\textwidth}|p{0.31\textwidth}|}\hline
\multicolumn{1}{|c|}{Query Sentence} & \multicolumn{1}{|c|}{Semantically Similar} & \multicolumn{1}{|c|}{Syntactically Similar} \\\hline
with a gun , you feel more dangerous . i would n't know . & i 'll give him a gun if it makes you feel any better . & all these days , i could n't work , i could n't sleep . \\\hline
if you want the loo , it 's just out here in the corridor . & the bathroom 's at the end of the corridor , if you need it . & when you need a car , you pull it right out of the tree . \\\hline
it 's bumpy , but we 'll be past it in a few minutes . & it 's gon na be a little bumpy till we get over the mountains . & it looks negligible , but it will be difficult to get any useful samples . \\\hline
`` who are you to talk to me like that ? & talk to me , just tell me who you are . & what 're you gon na do to me with that ? \\\hline
十五 步 之 远 我们 的 客厅 是 屋子 里 最 大 的 房间 & 这 间 客房 是 更 大 的 ， 我 需要 多 一点 空间 。 & 如今 整个 银河 中 你们 的 星球 是 文明 程度 最 低 的 星球 \\
(fifteen step away our living room be house inside the biggest room) & (this guest room be bigger , I need more a little space .) & (now entire Galaxy in you of planet be civilization degree the most low planet) \\\hline
你 从 窗户 外 可以 看到 它 或在 电视 上 也 会 看到 它 & 你 几乎 不 可能 再 近 距离 看到 它的 ， 用 望远镜 也 不行 & 我 在 国会 上 讲 过 我们 需要 建桥 但 我们 会 靠 自己 \\
(you from window outside can see it or in tv also can see it) & (you almost no possible again close distance see it , use telescope also not work) & (i at congress talked we need build a bridge but we can depend on ourselves) \\\hline
我们 完全 有 能力 为 我们 的 女儿 举办 一 个 独特 的 婚礼 & 我们 糊弄 他们 , 我们 将 会 举行 一 个 很 大 的 婚礼 & 我们 很 可能 会 为 我们 的 驻罗 分公司 选 一些 管理层 的 人士 \\
(we absolutely have capability for our daughter hold a unique wedding) & (we fool them , we will host a very big wedding) & (we very possible can for our zhuoluo branch select some management person) \\\hline
看 它 的 胸腔 在 动 他 正在 呼吸 大量 氧气 & Foreman 做 了 胸腔 穿刺术 来 排出 她 肺部 的 积水 & 闯进 她 的 公寓 拿着 刀 你 已经 越线 太 远了 \\
(look its chest cavity be move he being breathe great amount oxygen) & (Forema did chest cavity puncture to discharge she lungs stagnant water) & (break into she apartment hold knife you already cross the line too far) \\\hline
\end{tabular}
    \caption{The most similar sentences to particular query sentences in terms of the semantic variable and syntactic variable based on the model trained on Zh-En bitext. Each Chinese sentence is followed by a gloss. More examples are in the supplementary material.}
    \vspace{-1em}
    \label{tab:near-neighbor-analysis-mono2}
\end{table*}
\end{CJK*}

We show the nearest neighbour examples in Table \ref{tab:near-neighbor-analysis-mono2}.

\section{Runtime and Computing Infrastructures}

Our models are trained on machines equipped with a single GPU, such as NVIDIA TITAN X or NVIDIA 2080 Ti. We train all of our models for 20 epochs. For the big model, it takes approximately 11.67 hours to finish one epoch. For other mVGVAE models, it takes 6.67 hours to finish one epoch.

\section{Human Evaluation}
\begin{table}[t]
    \centering
    \begin{tabular}{|p{0.4\textwidth}|}\hline
1 = The two sentences are completely dissimilar in structure. \\\hline
        2 = The two sentences do not have very similar structure overall, but there are some similarities (e.g., both start with an independent clause that begins with a subject, or both contain a dependent clause introduced by “that”). \\\hline
        3 = The two sentences are roughly equivalent in overall structure, but individual clauses in the sentences have different structures. \\\hline
        4 = The two sentences have similar structure overall, but there are some small differences (e.g., one sentence has more modifiers (adjectives or adverbs) than the other). \\\hline
        5 = The two sentences are completely equivalent, as they have the same sentence structure.\\\hline
    \end{tabular}
    \caption{Detailed explanation for each option for human evaluations.}
    \label{sup:tab:human_evaluation_options}
\end{table}
During human evaluations, annotators were asked one question: "On a scale of 1-5, how much do you think the structure of sentence1 and the structure of sentence2 are similar?", and we show the detailed explanation for each option in Table \ref{sup:tab:human_evaluation_options}.

\section{Controlled Paraphrase Generation}

\begin{table*}[t]
    \centering
    \small
\begin{tabular}{|c|c|c|c|c|c|c|c|}\hline
& BLEU & ROUGE-1 & ROUGE-2 & ROUGE-L & METEOR & ST-r & ST-s \\\hline
\multicolumn{8}{|c|}{Return-input baselines}\\\hline
Syntactic input & 3.3 & 24.4 & 7.5 & 29.1 & 12.1 & 5.9 & 0.0 \\
Semantic input & 18.5 & 50.6 & 23.2 & 47.7 & 28.8 & 12.0 & 13.0 \\\hline
\multicolumn{8}{|c|}{Models trained on the ParaNMT dataset}\\\hline
Prior work (En-En) 
& 13.6 & 44.7 & 21.0 & 48.3 & 24.8 & 6.7 & 3.4 \\
En-En & 13.0 & 44.0 & 20.0 & 47.4 & 23.6 & 6.6 & 3.3 \\\hline
\multicolumn{8}{|c|}{Models trained on bitext (our work)}\\\hline
Cs-En & 12.1 & 41.5 & 18.0 & 45.1 & 22.4 & 6.9 & 3.4 \\\hline
Zh-En & 11.5 & 42.7 & 18.3 & 46.1 & 22.2 & 6.7 & 3.4 \\
Zh-En Big & 12.5 & 44.6 & 18.7 & 47.4 & 23.2 & 6.7 & 3.5 \\\hline
De-En & 12.6 & 42.1 & 18.4 & 45.8 & 22.1 & 6.6 & 3.2 \\\hline
Es-En & 11.7 & 41.3 & 17.5 & 44.9 & 21.8 & 6.8 & 3.7 \\\hline
Fr-En & 11.3 & 41.1 & 17.6 & 44.6 & 21.2 & 6.7 & 3.3 \\\hline
\end{tabular}
    \caption{Test set results for English controlled paraphrase generation. The models are trained on 5 million sentence pairs. Lower is better for ST-r and ST-s.}
    \vspace{-1em}
    \label{sup:tab:en-para-gen}
\end{table*}

\begin{table*}[t]
    \centering
    \small
\begin{tabular}{|c|c|c|c|c|c|c|c|}\hline
& BLEU & ROUGE-1 & ROUGE-2 & ROUGE-L & METEOR & ST-r & ST-s \\\hline
\multicolumn{8}{|c|}{Return-input baselines}\\\hline
Syntactic input & 6.2 & 29.9 & 9.2 & 34.4 & 13.2 & 13.9 & 0.0 \\
Semantic input & 49.0 & 77.4 & 57.0 & 65.4 & 42.5 & 18.7 & 22.4 \\\hline
\multicolumn{8}{|c|}{Our work}\\\hline
Zh-En & 12.8 & 46.1 & 20.6 & 46.4 & 21.2 & 15.8 & 12.6 \\
Zh-En Big & 16.6 & 52.4 & 26.0 & 50.8 & 24.8 & 15.5 & 12.0 \\\hline
\end{tabular}
    \caption{Test set results for the Chinese controlled paraphrase generation. Lower is better for ST-r and ST-s.}
    \vspace{-1em}
    \label{sup:tab:zh-para-gen}
\end{table*}

\begin{table*}[t]
    \centering
    \small
\begin{tabular}{|c|c|c|c|c|c|c|c|}\hline
& BLEU & ROUGE-1 & ROUGE-2 & ROUGE-L & METEOR & ST-r & ST-s \\\hline
\multicolumn{8}{|c|}{Return-input baselines}\\\hline
Syntactic input & 3.3 & 24.4 & 7.5 & 29.1 & 12.2 & 5.9 & 0.0 \\\hline
\multicolumn{8}{|c|}{Neural machine translation baselines}\\\hline
OpenNMT & 11.0 & 42.4 & 17.0 & 43.1 & 23.4 & 11.5 & 12.0 \\
Google Translate & 14.5 & 47.7 & 20.8 & 46.5 & 27.3 & 11.6 & 12.6  \\\hline
\multicolumn{8}{|c|}{Our work}\\\hline
Zh-En & 10.9 & 40.6 & 17.9 & 44.2 & 21.3 & 6.7 & 3.3 \\
Zh-En Big & 12.1 & 42.2 & 18.3 & 45.6 & 22.3 & 6.6 & 3.3 \\\hline
\end{tabular}
    \caption{Test set performance for Chinese-to-English controlled machine translation. Lower is better for ST-r and ST-s.}
    \vspace{-1em}
    \label{sup:tab:zh2en-para-gen}
\end{table*}

\begin{table*}[t]
    \centering
    \small
\begin{tabular}{|c|c|c|c|c|c|c|c|}\hline
& BLEU & ROUGE-1 & ROUGE-2 & ROUGE-L & METEOR & ST-r & ST-s \\\hline
\multicolumn{8}{|c|}{Return-input baselines}\\\hline
Syntactic input & 6.2 & 29.9 & 9.2 & 34.4 & 13.2 & 13.9 & 0.0 \\\hline
\multicolumn{8}{|c|}{Neural machine translation baselines}\\\hline
OpenNMT & 12.3 & 40.4 & 20.0 & 39.9 & 17.7 & 19.2 & 20.4 \\
Google Translate & 30.3 & 63.2 & 37.7 & 56.6 & 32.0 & 18.5 & 21.3 \\\hline
\multicolumn{8}{|c|}{Our work}\\\hline
Zh-En & 9.6 & 40.2 & 16.1 & 41.4 & 18.1 & 16.1 & 11.0 \\
Zh-En Big & 11.9 & 44.3 & 19.1 & 44.5 & 20.4 & 15.7 & 10.0 \\\hline
\end{tabular}
    \caption{Test set performance for English-to-Chinese controlled machine translation. Lower is better for ST-r and ST-s.}
    \label{sup:tab:en2zh-para-gen}
\end{table*}

We report BLEU, ROUGE-1, ROUGE-2, ROUGE-L, and METEOR. For ST metrics, we report ST-r and ST-s, which are computed between the generations and references, and the generations and syntactic inputs respectively.

We report English controlled paraphrase generation results in Table \ref{sup:tab:en-para-gen}. It is interesting to see that models trained on other language pairs can also lean to perform this task.

We report results for Chinese controlled paraphrase generation in Table \ref{sup:tab:zh-para-gen}, Chinese-to-English controlled machine translation in Table \ref{sup:tab:zh2en-para-gen}, and English-to-Chinese controlled machine translation in Table \ref{sup:tab:en2zh-para-gen}. \footnote{When computing ROUGE scores for Chinese, we map Chinese characters to unique IDs to avoid encoding problems.}

\section{Semantic Disentanglement}
We report results for semantic evaluation for other languages in Table \ref{sup:tab:en-sts-res} and Table \ref{sup:tab:mul-sts-res}. For Spanish, we use the Spanish test set from STS 2017~\cite{agirre-etal-2014-semeval}. We found that models trained on other language pairs also learn to disentangle latent representations.

\begin{table*}[t]
    \centering
    \small
\begin{tabular}{|c|c|c|c|c|c|c|c|c|c|c|c|c|c|c|c|}
\hline
\multirow{2}{*}{ } & \multicolumn{3}{c|}{ En-En } & \multicolumn{3}{c|}{ Fr-En }  & \multicolumn{3}{c|}{ De-En } & \multicolumn{3}{c|}{ Es-En } & \multicolumn{3}{c|}{ Zh-En } \\
\cline{2-16}
& sem. & syn. & $\Delta$ & sem. & syn. & $\Delta$ & sem. & syn. & $\Delta$ & sem. & syn. & $\Delta$ & sem. & syn. & $\Delta$ \\\hline
Prior work & 74.3 & 7.4 & 66.9 & - & - & - & - & - & - & - & - & & - & - & - \\\hline
Our work & 74.9 & 12.0 & 62.9 & 73.2 & 10.4 & 62.8 & 68.6 & 12.1 & 56.5 & 72.6 & 12.9 & 59.7 & 73.6 & 16.3 & 57.3 \\\hline
\end{tabular}
    \caption{Pearson's correlation (\%) for the English semantic similarity task.}
    \label{sup:tab:en-sts-res}
\end{table*}

\begin{table}[t]
    \centering
    \small
\begin{tabular}{|c|c|c|c|c|c|c|}
\hline
\multirow{2}{*}{ } & \multicolumn{3}{c|}{ Es-En } & \multicolumn{3}{c|}{ Zh-En } \\\cline{2-7}
& sem. & syn. & $\Delta$ & sem. & syn. & $\Delta$ \\\hline
Fasttext & \multicolumn{3}{|c|}{ 48.0 } & \multicolumn{3}{c|}{ 74.0 }\\\hline
Our work & 76.3 & 49.0 & 27.3 & 76.2 & 23.5 & 52.8 \\\hline
\end{tabular}
    \caption{Pearson's correlation (\%) for the Spanish semantic similarity and Chinese semantic similarity tasks.}
    \label{sup:tab:mul-sts-res}
\end{table}

\section{Syntactic Disentanglement}

We report results for syntactic evaluation for other languages in Table \ref{sup:tab:en-syn-res} and Table \ref{sup:tab:mul-syn-res}. We found that models trained on other language pairs also learn to disentangle latent representations.

\begin{table}[t]
    \centering
    \small
\begin{tabular}{|c|c|c|c|c|c|c|}
\hline
\multirow{2}{*}{ } & \multicolumn{3}{c|}{ POS Accuracy } & \multicolumn{3}{c|}{ CP Labeled $F_1$ } \\\cline{2-7}
& sem. & syn. & $\Delta$ & sem. & syn. & $\Delta$ \\\hline
Oracle & \multicolumn{3}{c|}{62.3} & \multicolumn{3}{c|}{71.1} \\
Random & \multicolumn{3}{c|}{12.9} & \multicolumn{3}{c|}{19.2} \\\hline
Prior work & 20.3 & 43.7 & 23.4 & 24.8 & 40.9 & 16.1 \\\hline
En-En & 19.6 & 44.9 & 25.3 & 24.2 & 42.4 & 18.2 \\\hline
Fr-En & 19.9 & 41.8 & 21.9 & 24.2 & 41.8 & 17.6 \\\hline
De-En & 20.1 & 44.7 & 24.6 & 24.4 & 42.0 & 17.6 \\\hline
Es-En & 19.7 & 44.0 & 24.4 & 24.1 & 41.5 & 17.4 \\\hline
Zh-En & 19.7 & 44.4 & 24.7 & 24.4 & 41.7 & 17.3 \\\hline
\end{tabular}
    \caption{POS tagging accuracy (\%) and constituent parsing labeled $F_1$ scores (\%) for the English syntactic evaluation. The models are trained on 5 million sentence pairs.}
    \label{sup:tab:en-syn-res}
\end{table}

\begin{table}[t]
    \centering
    \small
\begin{tabular}{|c|c|c|c|c|c|c|}
\hline
\multirow{2}{*}{ } & \multicolumn{3}{c|}{ POS Accuracy } & \multicolumn{3}{c|}{ CP Labeled $F_1$ } \\\cline{2-7}
& sem. & syn. & $\Delta$ & sem. & syn. & $\Delta$ \\\hline
Oracle Fr & \multicolumn{3}{c|}{ 71.7 } & \multicolumn{3}{c|}{ 92.8 } \\
Random Fr & \multicolumn{3}{c|}{ 18.2 } & \multicolumn{3}{c|}{ 31.0 } \\\cline{2-7}
fastText & \multicolumn{3}{c|}{ 28.5 } & \multicolumn{3}{c|}{ 34.7 } \\\cline{2-7}
Fr-En & 22.5 & 51.2 & 28.7 & 33.1 & 42.5 & 9.4 \\\hline
Oracle De & \multicolumn{3}{c|}{ 67.7 } & \multicolumn{3}{c|}{ 99.9 } \\
Random De & \multicolumn{3}{c|}{ 17.7 } & \multicolumn{3}{c|}{ 46.5 } \\\cline{2-7}
fastText & \multicolumn{3}{c|}{ 27.0 } & \multicolumn{3}{c|}{ 51.7 } \\\cline{2-7}
De-En & 21.5 & 47.4 & 25.9 & 50.3 & 58.1 & 7.8 \\\hline
Oracle Es & \multicolumn{3}{c|}{ 56.8 } & \multicolumn{3}{c|}{ 64.2 } \\
Random Es & \multicolumn{3}{c|}{ 9.7 } & \multicolumn{3}{c|}{ 16.8 } \\\cline{2-7}
fastText & \multicolumn{3}{c|}{ 21.7 } & \multicolumn{3}{c|}{ 23.2 } \\\cline{2-7}
Es-En & 13.7 & 34.9 & 21.2 & 19.0 & 34.0 & 15.0 \\\hline
Oracle Zh & \multicolumn{3}{c|}{ 56.6 } & \multicolumn{3}{c|}{ 58.2 } \\
Random Zh & \multicolumn{3}{c|}{ 13.4 } & \multicolumn{3}{c|}{ 17.8 } \\\cline{2-7}
fastText & \multicolumn{3}{c|}{ 18.7 } & \multicolumn{3}{c|}{21.7 } \\\cline{2-7}
Zh-En & 15.0 & 36.2 & 21.2 & 19.3 & 32.3 & 12.9 \\\hline
\end{tabular}
    \caption{POS tagging accuracy (\%) and constituent parsing labeled $F_1$ scores (\%) for the multilingual syntactic evaluation. The models are trained on 5 million sentence pairs.}
    \label{sup:tab:mul-syn-res}
\end{table}

\end{document}